\documentclass[journal]{IEEEtran}
\usepackage[graphicx]{realboxes}
\usepackage{xcolor,soul,framed} 
\usepackage{cite}
\usepackage{multirow}
\usepackage{fontawesome}
\colorlet{shadecolor}{yellow}
\usepackage{caption}
\usepackage{subcaption}
\usepackage[cmex10]{amsmath}
\usepackage[colorinlistoftodos]{todonotes}
\usepackage{comment}
\usepackage{enumerate}

\newcommand\textblue[1]{\textcolor{blue}{#1}}
\newcommand\textred[1]{\textcolor{red}{#1}}
\newcommand\quickthings[1]{\textblue{\\\faQuestion #1}}
\newcommand\maybelater[1]{\textred{\\\faClockO#1}}
\usepackage{multirow}
\usepackage{colortbl}
\usepackage{siunitx}
\usepackage{booktabs}
\definecolor{lightgray}{gray}{0.9}

\renewcommand\quickthings[1]{}
\renewcommand\maybelater[1]{}
\usepackage{bm}
\usepackage[linesnumbered,ruled,vlined]{algorithm2e}
\begin{document}
\bstctlcite{IEEEexample:BSTcontrol}
\title{The Hidden Adversarial Vulnerabilities of Medical Federated Learning}
\author{
    Erfan~Darzi$^{1}$, 
    Florian Dubost$^{2}$, 
    Nanna. M. Sijtsema$^{3}$, 
    P.M.A van Ooijen$^{3}$\\
    \textit{$^{1}$Harvard University, Boston, MA, USA}\\
    \textit{$^{2}$Google, Mountain View, CA, USA}\\
    \textit{$^{3}$Department of Radiotherapy, University Medical Center Groningen, \\University of Groningen, Groningen, The Netherlands}
}
\maketitle

\begin{abstract}
In this paper, we delve into the susceptibility of federated medical image analysis systems to adversarial attacks. Our analysis uncovers a novel exploitation avenue: using gradient information from prior global model updates, adversaries can enhance the efficiency and transferability of their attacks. Specifically, we demonstrate that single-step attacks (e.g. FGSM), when aptly initialized, can outperform the efficiency of their iterative counterparts but with reduced computational demand. Our findings underscore the need to revisit our understanding of AI security in federated healthcare settings.
\end{abstract}

\section{Introduction}
Federated learning represents a novel shift in machine learning, advocating for a decentralized method of training models across multiple servers or devices without the need to centralize data. This is particularly significant in areas like healthcare where sharing patient data across different entities can breach privacy regulations\cite{darzidehkalani2023comparative}. By keeping the data local, federated learning not only safeguards patient privacy but also facilitates collaborative model training without the transfer of sensitive medical details.

With the rise of AI applications in healthcare, there's been a concurrent surge in research focusing on adversarial attacks. These are tactics where attackers make slight alterations to the inputs of neural networks, leading them to produce erroneous outputs. Evaluating the security of AI in healthcare, particularly in decentralized settings, has become paramount, and many researchers are delving into this arena.

While the security of AI systems, especially in healthcare, has gained traction due to its widespread deployment and the inherent need to ensure its safety in practical applications, there is still a lot to be explored. Some studies have probed adversarial attacks in the healthcare setting, but many are limited to centralized systems or rely solely on white-box models. Amongst these, a few have assessed the susceptibility of healthcare data to adversarial threats under various conditions, including centralized, black-box, and white-box scenarios. Interestingly, many of these studies underscore the vulnerability of these systems to conventional adversarial attacks, often presuming a black-box setting, which suggests that the adversary doesn't have complete access to the global model or the other participants.

In a federated setting, where adversaries might participate in the federated learning process as a client, the potential for heightened risks, even with limited information, could challenge our current understanding of security. This introduces an important question: Does federated learning unintentionally reveal more vulnerabilities?

\subsection*{Our Contributions}

In this paper, we examine the susceptibility of federated medical image analysis systems to adversarial attacks. We demonstrate how an adversary can exploit the unique characteristics of a federated environment, using the distributed nature of federated learning and additional information at their disposal, to enhance the transferability of their attacks.

Our primary contributions include:

\textbf{Highlighting Vulnerabilities}: We evaluate the potential vulnerabilities of federated learning deployments, particularly within medical imaging, and provide an assessment of attack efficiency, attack transferability, and parameter dependency.

\textbf{Efficiency of Single-Step Attacks} Our research reveals that, given the right initialization, an adversary can use single-step attacks that are potentially more effective than multi-step iterative approaches with higher computational demands. This effectiveness is amplified when the adversary utilizes information from prior global model updates.

\textbf{Enhanced Transferability} We present evidence that adversarial examples, when generated using information from a federated environment, can result in a higher success rate for the attacker. Such examples are more transferable than those produced by conventional black box techniques. This is particularly crucial in a medical context, where an adversarial client could craft adversarial samples capable of misleading other clients to a significant extent.

\section{Background and Related Works}

Various studies have explored adversarial attacks within the realm of medical imaging, delving into their improvement and evaluation. However, there is a noticeable gap in research dedicated to evaluating these attacks in federated settings. In this section, we summarize relevant concepts regarding federated learning and adversarial attacks, specifically within the context of medical imaging.

\textbf{Federated Learning:} Federated Learning (FL) is a methodology that enables decentralized training of machine learning models over multiple devices or servers, without relocating the original data. The core mechanism of FL can be articulated as:

Given \(N\) clients, each having its dataset \(D_i\) where \(i \in \{1, ..., N\}\), every client determines updates using its localized dataset and then forwards these updates to a central server.

For each iteration \(t\):
\\1. The server picks a subset \(S_t\) of \(m\) clients.
\\2. Every client \(i \in S_t\) calculates its local update as:
   \[ w_{t+1}^i = w_t^i - \eta \nabla L_i(w_t^i) \]
   Here, \(L_i\) denotes the local loss based on client \(i\)'s data, \(w_t^i\) represents the model parameters for client \(i\) during iteration \(t\), and \(\eta\) is the learning rate.
\\3. The server consolidates the updates:
   \[ w_{t+1} = \sum_{i \in S_t} \frac{|D_i|}{|D|} w_{t+1}^i \]
   In this equation, \(|D|\) is the cumulative number of samples across all clients, and \(|D_i|\) refers to the samples held by client \(i\).






\textbf{Adversarial Attacks:} The objective of adversarial attacks is to subtly alter input data, often in barely detectable ways, to deceive machine learning  models. Prominent adversarial attack strategies in centralized settings encompass the Fast Gradient Sign Method (FGSM) and Projected Gradient Descent (PGD). For a model \(f\) with parameters \(\theta\), a loss function \(L\), and an input-output pair \((x, y)\), FGSM's adversarial disruption is:
\begin{equation}
\delta = \epsilon \cdot \text{sign}(\nabla_x L(f_\theta(x), y))
\end{equation}

Here, \(\epsilon\) dictates the perturbation's intensity. PGD is often likened to an iterative version of FGSM. This method creates adversarial examples by iteratively adjusting the input data. For each iteration, the PGD attack computes the gradient of the loss concerning the input data, makes a small step in the gradient direction, and then projects the resultant adversarial example back into the allowable \(\epsilon\)-boundary to ensure perturbations do not exceed a pre-defined magnitude. The iterative update for PGD can be described as:
\begin{equation}
x' = \text{Clip}_{x-\epsilon, x+\epsilon}\left(x + \alpha \times \text{sign}(\nabla_x L(f_\theta(x), y))\right)
\end{equation}
In the equation above, \(\alpha\) denotes the step size of each PGD iteration, and the \(\text{Clip}_{a, b}(z)\) function ensures that every element of \(z\) lies within the range [a, b]. 

Attack strategies can be broadly categorized based on their level of knowledge about the target model.

- White-box Attacks: In this category, attackers have comprehensive knowledge of the target model, encompassing its architecture, parameters, and the data on which it was trained. Techniques such as FGSM and PGD, when executed with full model access by the attacker, are considered white-box attacks.

- Black-box Attacks: These attacks are characterized by limited insight into the model. While attackers might be familiar with the model's architecture, they lack knowledge of its specific parameters and training data. A key aspect of black-box attacks is their reliance on the transferability of adversarial examples, meaning perturbations crafted for one model can often mislead another.

\textbf{Adversarial Attacks on Medical Images:} The susceptibility of deep learning models in medical imaging to adversarial attacks has been an increasing area of interest in recent research. Numerous studies have surfaced, with each addressing different facets of adversarial vulnerabilities specific to medical imaging. 

Raman et al.\cite{bms2022analysis} investigated gradient-based adversarial attacks such as FGSM, MI-FGSM, BIM, and PGD, specifically in the realm of medical image classification. Ma et al.\cite{ma2021understanding} offered an in-depth analysis of adversarial examples in medical DNNs, emphasizing the \(L_{\infty}\) perturbation constraint and primarily focusing on untargeted white-box attacks. Bortsova et al.\cite{bortsova2021adversarial} expanded the conversation to include black-box attacks, a noteworthy extension since black-box scenarios frequently mirror real-world situations more accurately. Their research examined crucial elements like weight initialization and variations in training data, addressing certain gaps not covered by earlier studies. Meanwhile, Hirano et al.\cite{hirano2021universal} shed light on Universal adversarial perturbations within the medical imaging sphere, further enhancing the comprehension of black-box scenarios.

It's important to note that many studies, including the one by Ma et al.\cite{ma2021understanding}, largely center around theoretical discussions. This focus may inadvertently neglect practical considerations, such as computational constraints and efficiency. While there's an extensive analysis available, the distinct challenges posed by federated learning environments and their related adversarial vulnerabilities are still underrepresented in the current literature.


\IEEEpeerreviewmaketitle

\section{Methodology}

While our previous work has shown that adversarial attacks have proven effective in federated settings and black-box environments\cite{darzi2023exploring}, the process for generating conventionally trained adversarial samples often overlooks the information available via the attack surface presented by federated learning. Indeed, executing adversarial attacks in a federated environment can be computationally expensive, requiring robust GPU computations, especially for iterative attacks. In this section, we demonstrate how an adversary can leverage previous model updates to execute more precise attacks. Empirically, we illustrate that the information embedded in model updates can be invaluable for an adversary, enhancing both efficiency and the ability to deceive other clients.

\subsection{Cross-round Noise}

\begin{figure}[t!]
 \centering
 \includegraphics[width=0.50\textwidth]{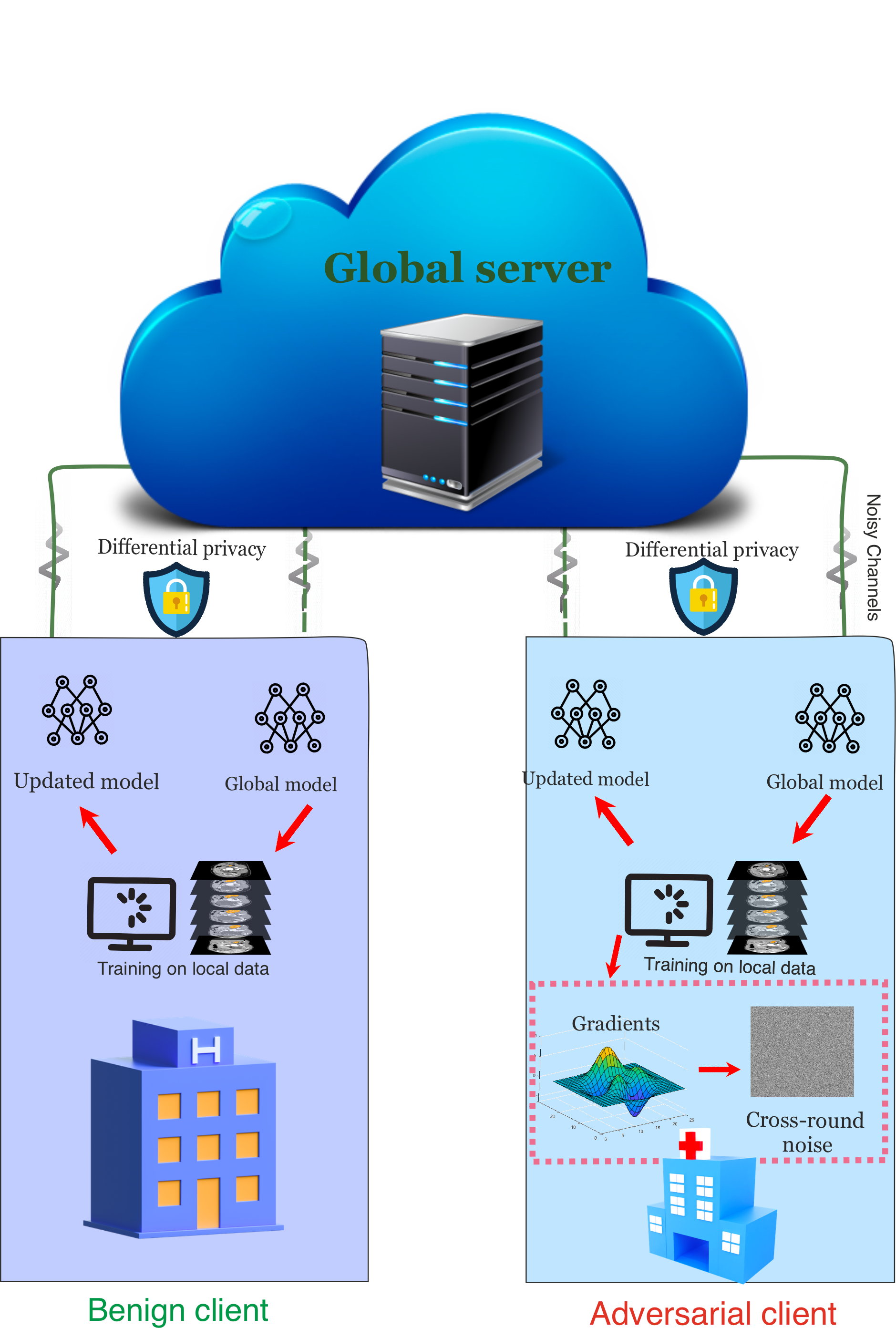}
 \caption{A schema of our method, at each federated round, the adversary transforms the gradients into cross-round noise, while acting as a benign client and not tampering the training process .}

 \label{fig:attack_hospitals}
\end{figure}

Given that adversarial samples are generated during inference, the adversary may lack the computational power typically available during training\cite{van2022ai}. Moreover, inference often takes place on third-party applications or edge devices without dedicated GPU support. Consequently, it becomes imperative for the adversary to craft attack samples with minimal computation. Iterative methods, like PGD, demand substantial computational resources to produce these samples, often exceeding the requirements of standard training \cite{zhang2019you}. This computational challenge intensifies when dealing with high-resolution or whole-slide images, as encountered in our pathology examples.

Additionally, for PGD to be effective, a close resemblance between the target and the adversary is essential to ensure transferability. Yet, introducing noise into models can lead to significant inter-client variability, serving as a potential protective measure.
By leveraging the pre-existing knowledge from the global model, attacks can be initialized from a more suitable baseline, potentially reducing computational overhead. 

We introduce a unique noise tensor called \textit{Cross-round noise}. This tensor is designed to extract features from the global model acquired during the federated learning process. Computed concurrently with local epochs, this noise tensor undergoes updates before transitioning to the subsequent federated round.

For generating adversarial noise, an adversary employs a specific function. This function determines noise based on the \(L_2\) regularized gradients during each federated training round. During testing, this pre-computed noise serves as the initial values for the adversarial attack algorithms.

\begin{algorithm}[t!]
\caption{Training procedure}
\label{alg:NbAFL}
\LinesNumbered
\KwData{$T$, $\beta$, ${w}^{(0)}$, $\mu$, $\epsilon$ and $\delta$}
{Initializing parameters: $t = 1$ and ${w}^{(0)}_{i} = {w}^{(0)}$ and ${{\mathcal{\delta}}^{(t)}=0}$} $\forall i,t$\\
\While {$t \le T$}
{
\textbf{Local training:}
\While {$\mathcal C_i\in \{\mathcal C_1, \mathcal C_2, \ldots,\mathcal C_{N}\}$}
{
Clients update their models ${w}^{(t)}_{i}$ as\\
\quad${w}^{(t)}_{i}=\arg\min\limits_{{w}_{i}}{\left(\mathcal{L}_{i}({w}_{i})\right)}$\\
Clients clip parameters ${w}^{(t)}_{i} = {w}^{(t)}_{i}/\max\left(1,\frac{\Vert{w}^{(t)}_{i}\Vert}{C}\right)$\\
Clients add Gaussian noise\\ $\widetilde{{w}}^{(t)}_{i}={w}^{(t)}_{i}+{n}^{(t)}_{i}$\\
}
\textbf{Global update:}

Global server individual models ${w}^{(t)}$ as\\
\quad\quad ${w}^{(t)} = \sum\limits_{i=1}^{N}{p_{i}\widetilde{{w}}^{(t)}_{i}}$\\
Global server adds Gaussian noise \\
\quad\quad$\widetilde{{w}_{i}}^{(t)}={w}^{(t)}+{n}_{\text i}^{(t)}$\\
\If{$t \ge \beta$}
    {
  \textbf{Adversarial client} $\mathcal C_{m},   \mathcal X \in \mathcal D_m:$
\\Adversary performs gradient regularization \\
\quad{${{\nabla_{\mathcal{X}}\mathcal{\hat{L}}\gets \nabla_{\mathcal{X}}\mathcal{L}(\mathcal {\delta}^{(t-1)},\widetilde{{w}_{m}}^{(t)}})}$}\\
Adversary projects the gradient
\quad$\mathcal{\delta}^{(t)}\gets \Pi_{P_\epsilon(\bm{0})}(\nabla_{\mathcal{X}}\mathcal{\hat{L})}$
}
$t\leftarrow t + 1$}
\KwResult{$\widetilde{{w}_{i}}^{(T)},{\mathbf{\mathcal{{\delta}}}^{T}}$}
\end{algorithm}

\subsubsection{Gradient calculation}

The noise is calculated by the gradients of model at each round. The adversary uses the loss:

\begin{equation}
\label{eqt:loss1}
 \nabla_{\bm{x}}{\mathcal{L}}(g(\bm{x+\delta}^{(t)};\bm{w^{(t)}}),g(\bm{x};\bm{w^{(t)}}))
\end{equation}
To update the stored values for the noise. In which $g(.;w^{t})$ refers to received global model in communication round $t$.  The adversary iteratively updates the noise by maximizing the loss between the model output for the noisy and the clean test data. 
The global model parameters change for the test dataset being the same in each round. 
Saving noise could be started after an arbitrary number of rounds ($\beta$) is passed and be calculated alongside local training with similar epochs. In this paper, we consider initializing from the last 10 (CRN10) and last 5 (CRN5) rounds. Fig. \ref{fig:attack_hospitals} shows a schema of our method.
\\\subsubsection{$L_{2}$ regularization}

Transferring gradient information between models might lead to drastic parameter change \cite{pan2019improving,dettmers20158}. Unlike some methods which manually reset the gradients, so they lose information \cite{zheng2020efficient}, we regularize the change in CRN by subtracting the mean value in each channel of the input noise. We apply a function for $L_{2}$ regularization the gradients.
\begin{equation}
    \nabla_{x_{i}}\mathcal{\hat{L}} = \nabla_{x_{i}}\mathcal{{L}} -  \frac{1}{M} \sum\limits_{j=1}^{M}\nabla_{x_{i,j}} \mathcal{{L}}
\end{equation}
The resulting value is channel-wise regularized loss, where $i$ indicates index of channel, and $j$ refers to individual pixels. This results in a smaller $L_{2}$ norm of gradient values, which is a proven measure against explosion or outlier gradients \cite{pascanu2013difficulty}.Then the final gradients are projected to a bounded $L_\infty$=$\epsilon$ norm around zero. 

\subsection{Inference phase attack}
Attacks are performed after training is done. FGSM and BIM use zero initialization, and PGD uses random initialization, then they compute the adversarial examples. Cross-round noise is added to the input test data as an initial point for these methods.

\section{Evaluation}

%

\subsection{Experimental setup}
\begin{figure*}[t!]
     \centering
     \begin{subfigure}[b]{0.99\textwidth}
         \centering
         \includegraphics[width=0.32\textwidth]{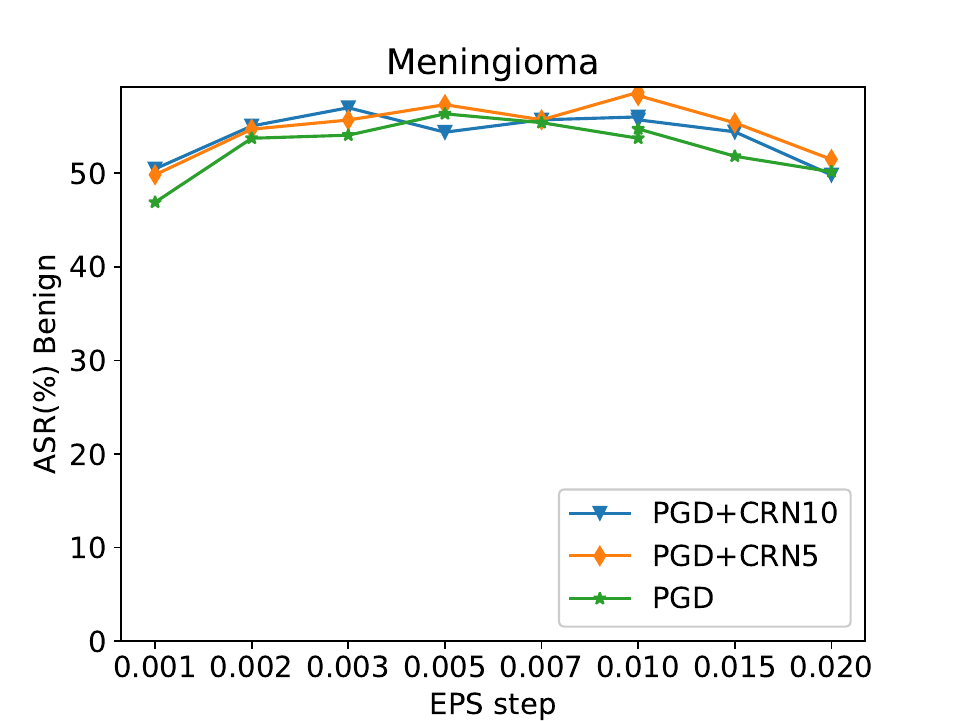}
          \includegraphics[width=0.32\textwidth]{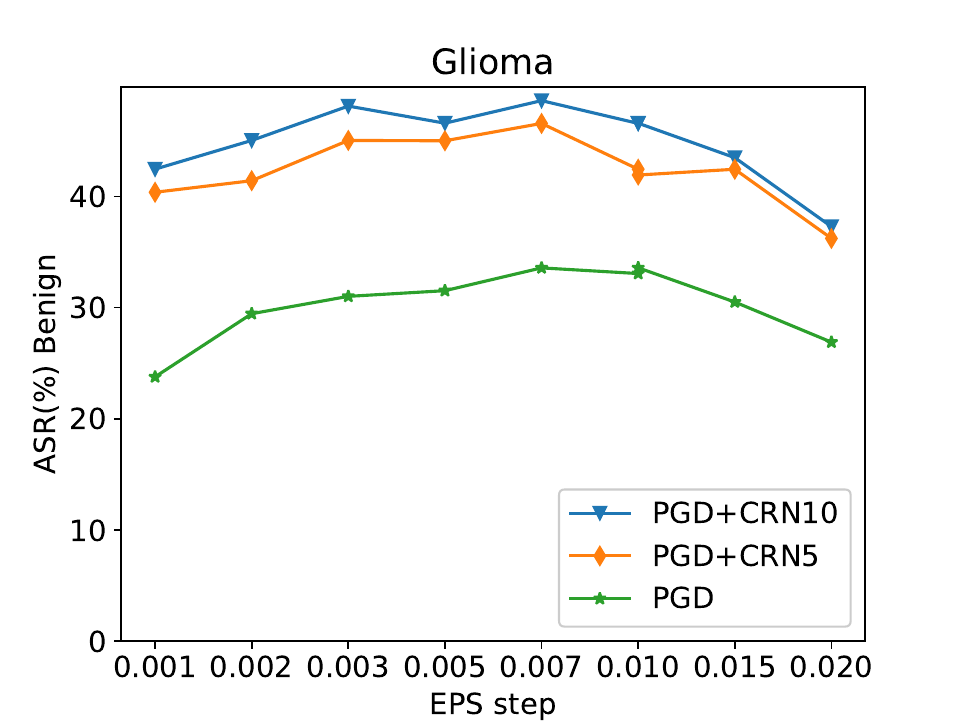}
           \includegraphics[width=0.32\textwidth]{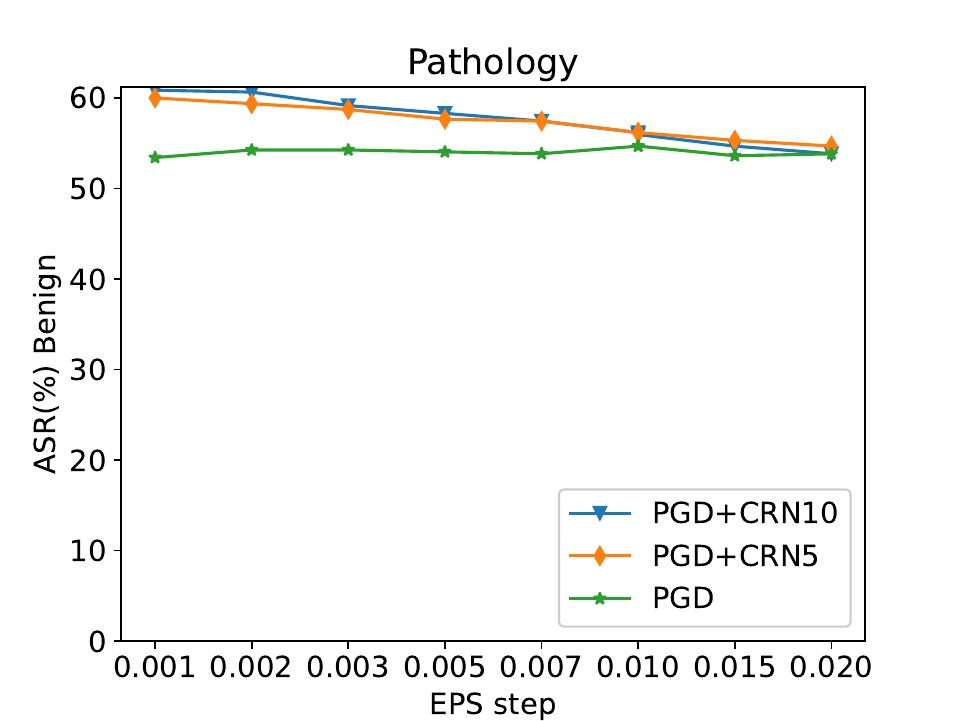}
         \caption{Average ASR on benign clients }
         \label{fig:y equals x}
     \end{subfigure}
     \hfill
     \begin{subfigure}[b]{0.99\textwidth}
          \centering
         \includegraphics[width=0.32\textwidth]{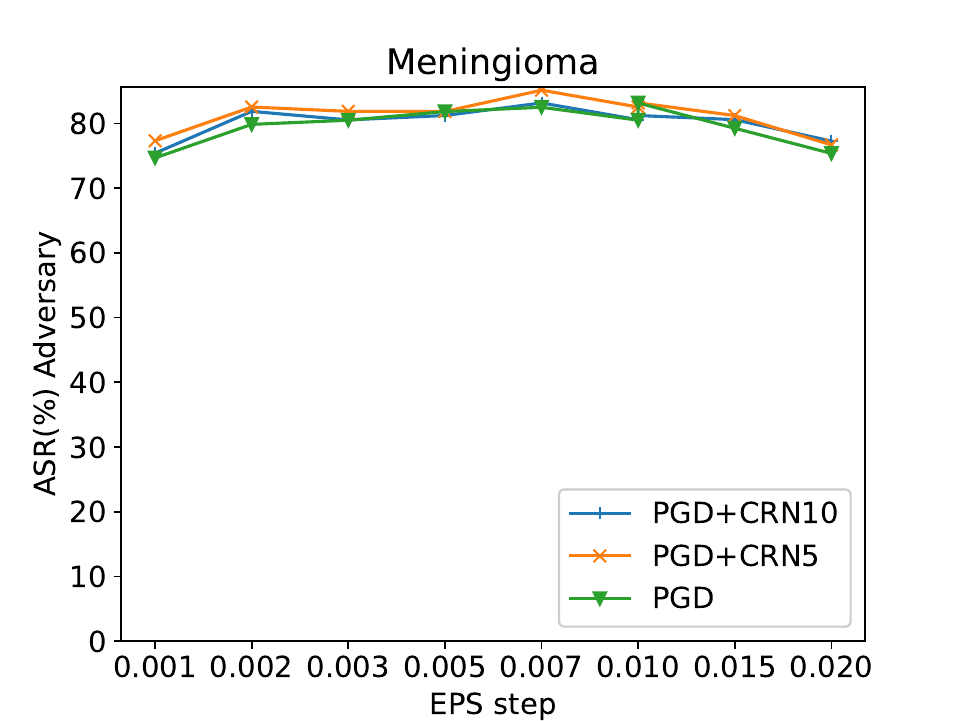}
          \includegraphics[width=0.32\textwidth]{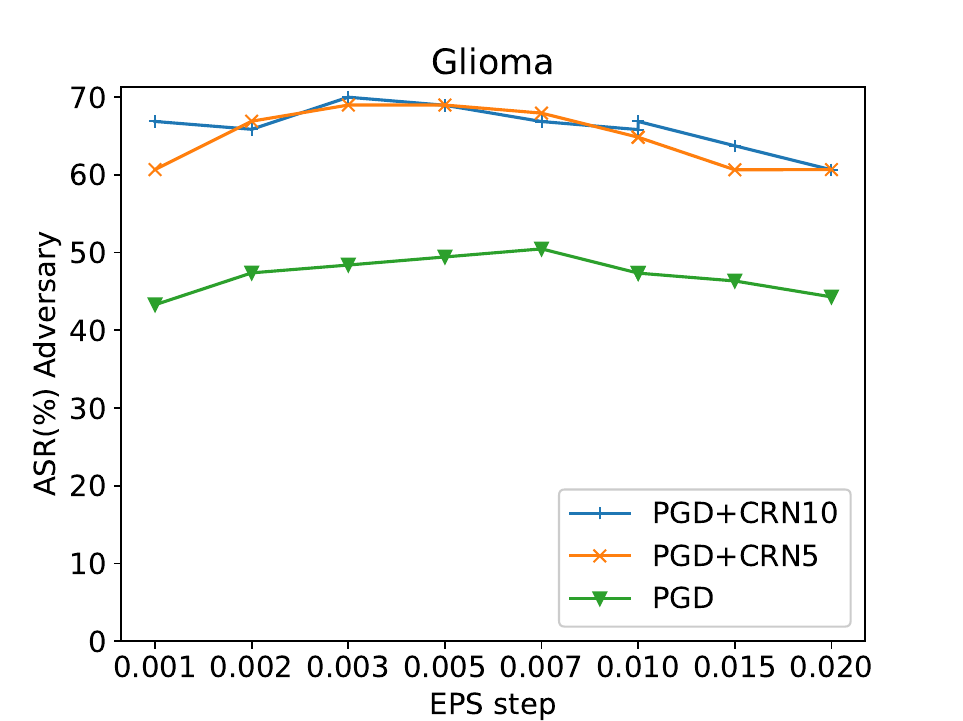}
           \includegraphics[width=0.32\textwidth]{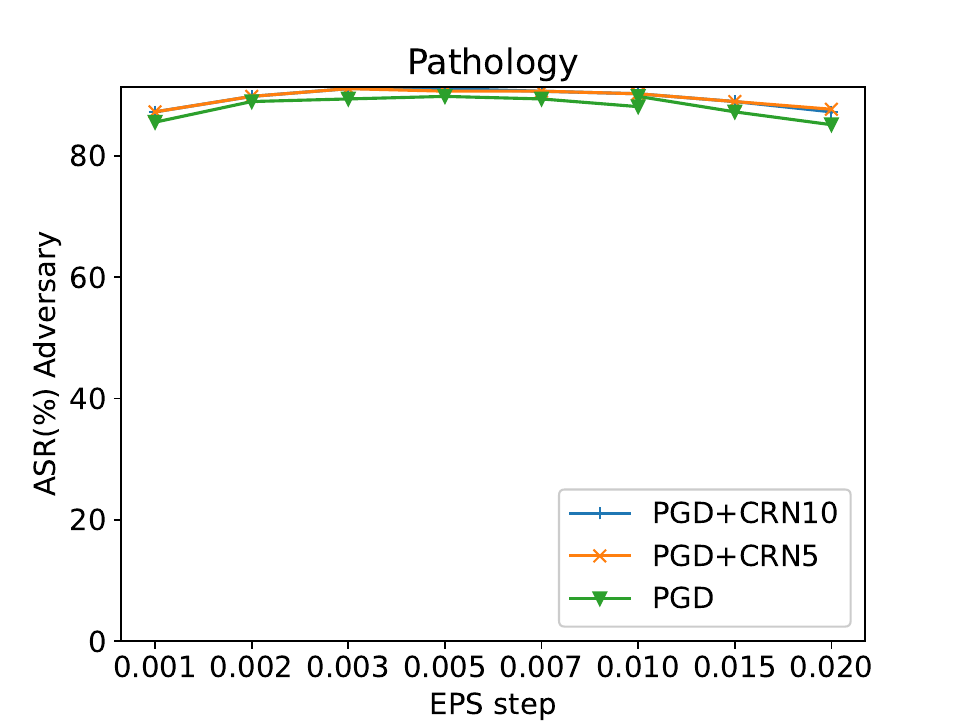}
         \caption{Average ASR on advesarial client }
         \label{fig:y equals x}
     \end{subfigure}
     \hfill
        \caption{Effect of Error perturbation step $\alpha$ on attack transferability. PGD attack with and without CRN initialization was performed. ASR is calculated on benign and adversary clients.The higher ASR on benign clients shows higher transferability}
        \label{fig:alpha}
\end{figure*}

\textbf{Datasets}
Our experiments are based on non-IID datasets, adhering to a setup similar to \cite{darzi2023exploring}. This approach ensures that each client receives a distinct dataset, mirroring the heterogeneity observed in real-world data distributions. We conducted our experiments on two primary datasets: 
\\\textit{Brain Cancer Detection}: This dataset, sourced from Kaggle\cite{sartaj}, consists of MRI images of the brain. These images are divided into three tumor categories and one for healthy instances. For the classifications of Meningioma and Glioma, 1437 and 1426 images were used, respectively.
\\\textit{Histopathologic Cancer Detection}: This dataset features images of metastatic lymph node tissues\cite{veeling2018rotation}, with each image having a resolution of $96 \times 96$ pixels. The main objective with these images is to classify tissue samples as either malignant or benign. Positive samples can be identified by a metastatic region centered within a $32\times32$ pixel area. Out of the entire collection, 2150 samples were labeled either as cancer or non-cancer and subsequently distributed among the clients.

\textbf{Preprocessing}
The images used for brain cancer detection were resized to a standard resolution of 100x100 pixels. They underwent a series of transformations, including rotation, flipping, and normalization. On the other hand, the images from the histopathologic cancer detection dataset went through transformations such as horizontal flipping and normalization. The dataset was divided with a train-test split of 62\%-38\%.

\textbf{Model Architectures} Our selected deep learning architecture is a Convolutional Neural Network (CNN) comprising six convolution layers followed by five fully connected layers. ReLU serves as the activation function with a dropout rate of 0.25. Before subjecting to adversarial attacks, all models were thoroughly trained. Each dataset utilized a classifier, trained using Cross-Entropy loss with SGD optimization. Three clients form the federated configuration, receiving data randomly with non-IID distributions. The FedAVG algorithm aggregates the models, with the aggregation proportioned by the training dataset's size. Each client undergoes training for 20 epochs per communication round, culminating in 50 federated rounds.
The adversarial role cycles between the three clients. Every client receives 100 test samples for each scenario to execute the attack.

\textbf{Evaluation metrics}
For the assessment, apart from the conventional clean accuracy (ACC), we deployed a variety of evaluative metrics. \textit{Attack Success Rate (ASR)} provides an analytical insight into the effectiveness of an adversary in modifying the model's predictive outcomes. The mathematical representation of ASR is given by:
\begin{equation}
    \text{ASR} = \frac{1}{N}\sum\limits_{i=1}^{N} (l_{\text{pre},i} \neq l_{\text{post},i})
\end{equation}
Here, \( l_{\text{pre},i} \) and \( l_{\text{post},i} \) symbolize the pre-attack and post-attack labels for the \( i^{th} \) sample. \( N \) stands for the cumulative count of images within the attack set. \textit{Average Attack Success Rate (AASR)} is essentially the mean derivative of the ASR, averaged over the entire client base. 
Another metric, the \textit{Average Error Transferability Rate (AETR)}, is used to evaluate transferability. With this metric, an adversary may attempt an attack by only selecting samples that successfully deceived its own model. AETR gauges the efficacy of these samples in changing the correct predictions of the target models. Unlike ASR, AETR excludes samples initially misclassified by the benign client's model\cite{papernot2017practical}. The formal definition of AETR is:
\begin{equation}
    \text{AETR} = \frac{1}{|\mathcal{C}|}\sum\limits_{c\in[\mathcal{C}]}\frac{1}{|\mathcal{D}_n|}\sum\limits_{i\in {D}_n} (l_{\text{pre}, i,c} \neq l_{\text{post}, i,c})
\end{equation}
where \( l_{\text{pre}, i,c} \) denotes the pre-attack label of the \( i^{th} \) sample for client \( c \), \( l_{\text{post}, i,c} \) represents the post-attack label of the \( i^{th} \) sample for client \( c \), and \( \mathcal{D}_n \) is a subset of the attack samples where the pre-attack labels match the target labels.

\subsection{Effect of Step Size ($\alpha$) on Attack Transferability}

We investigated the impact of the PGD attack's step size, $\alpha$, on the attack success rate across different models and datasets in a federated learning environment. Adjusting $\alpha$ can significantly influence the attack's potency and transferability. Specifically, the attack success rate was evaluated on both the benign clients and the adversary. 

Initially, the adversary executed the attack on its own model. Subsequently, the transferability of this attack to benign clients under varying step sizes ($\alpha$) was evaluated. Our results shown in Fig. \ref{fig:alpha} indicated that both extremely small and large $\alpha$ values led to suboptimal attack success rates, especially against benign clients. A step size of $\alpha = 0.005$ emerged as the most effective, consistently performing well with iterative PGD models, both in its pure form and when initialized with CRN5 and CRN10. The efficacy of $\alpha$ was found to be dataset-sensitive. For instance, the glioma dataset revealed a preference for mid-range $\alpha$ values in terms of higher attack success rates, while other datasets showed negligible differences.
\begin{table}[h!]
\centering
\setlength{\tabcolsep}{6pt}
\renewcommand\arraystretch{1.22}
\caption{ \small Comparison of iterative models and their computational efficiency, on performing computations on one batch of data.  ACC shows average client performance for unperturbed test data. AASR is average ASR on all clients.
}
\begin{tabular}{| *{5}{c|} }
\hline
Dataset  & Attack type & ACC & AASR & time (sec)
\\   \hline  
\multirow{3}{5em}{Meningioma}     &\textbf{PGD-1+CRN10}&\multirow{3}{3em}{84.12\%}&\textbf{63.92\%} & \textbf{0.217}  \\
 &PGD-20&&27.52\% & 3.423 \\
&PGD-40&&32.99\% & 6.794  \\ \hline
\multirow{3}{4em}{Pathology}     &\textbf{PGD-1+CRN10}&\multirow{3}{3em}{77.01\%}&\textbf{90.92\%} & \textbf{0.354} \\
&PGD-20&&60.98\% & 5.464\\
&PGD-40&&82.27\% & 10.843 \\ \hline
\multirow{3}{3em}{Glioma}     &\textbf{PGD-1+CRN10}&\multirow{3}{3em}{61.84\%}&\textbf{70.55}\% & \textbf{0.216}  \\
&PGD-20&&51.83\% &3.420  \\
&PGD-40&&63.77\% & 6.793  \\ \hline

\end{tabular}
\label{table_time} 
\end{table}

The addition of CRN initialization consistently boosted the attack success rate, particularly against benign clients. Nevertheless, the magnitude of this effect was dataset-dependent. Glioma, for instance, reaped the most significant benefits from CRN initialization. Also, unsurprisingly, the ASR was high on the adversary across all scenarios. However, benign clients were still susceptible, being deceived around 40\% to 60\% of the time.

\begin{figure*}[t!]
     \centering
     \begin{subfigure}[b]{1.05\textwidth}
         \centering
         \includegraphics[width=0.32\textwidth]{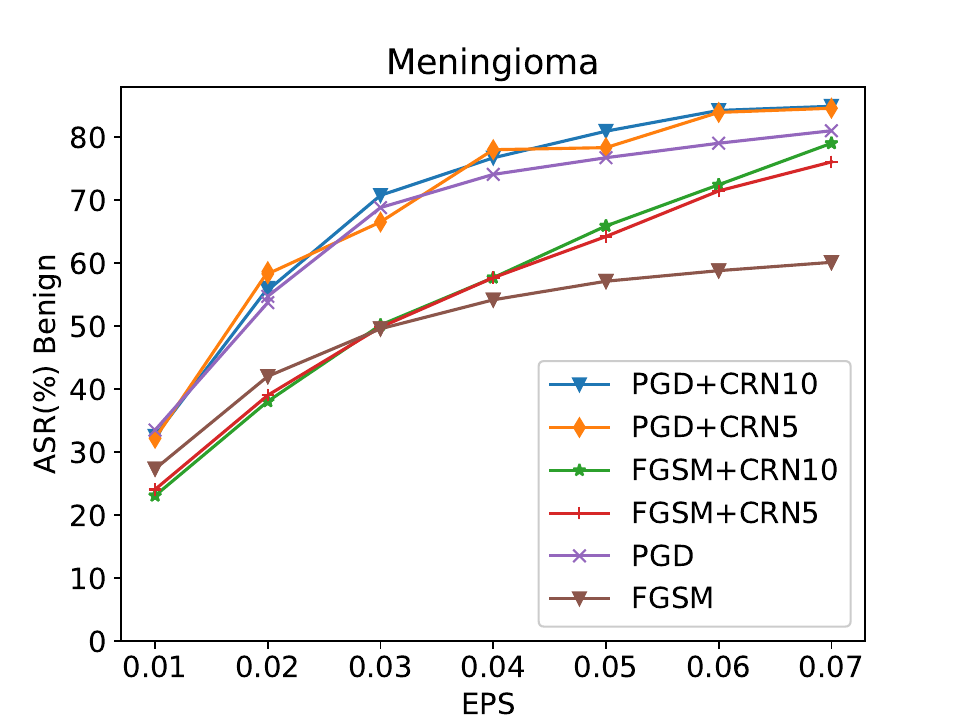}
          \includegraphics[width=0.32\textwidth]{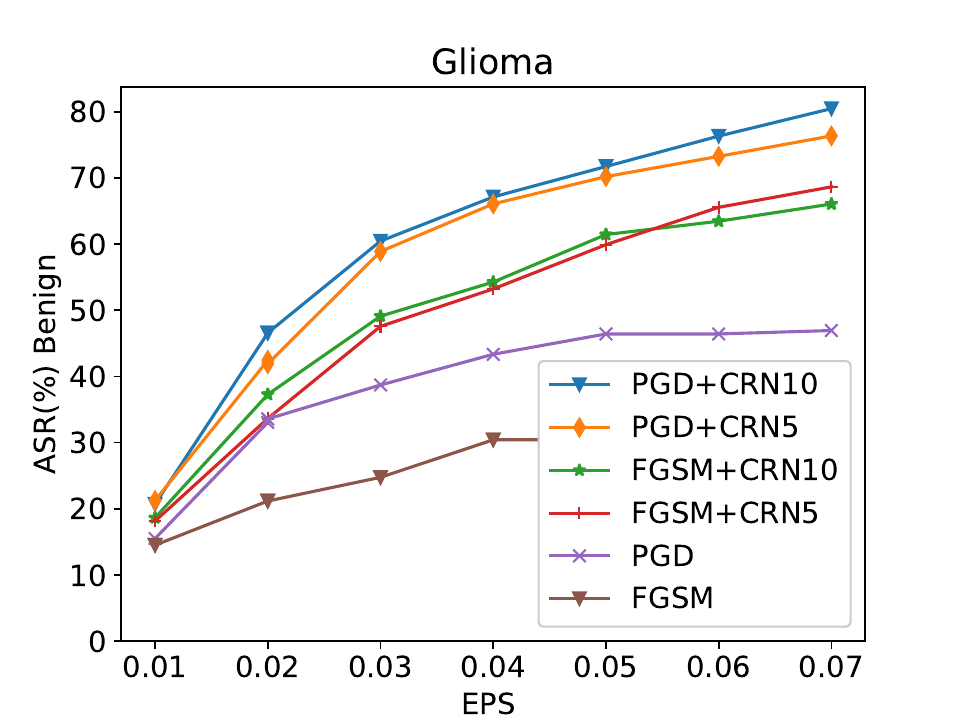}
           \includegraphics[width=0.32\textwidth]{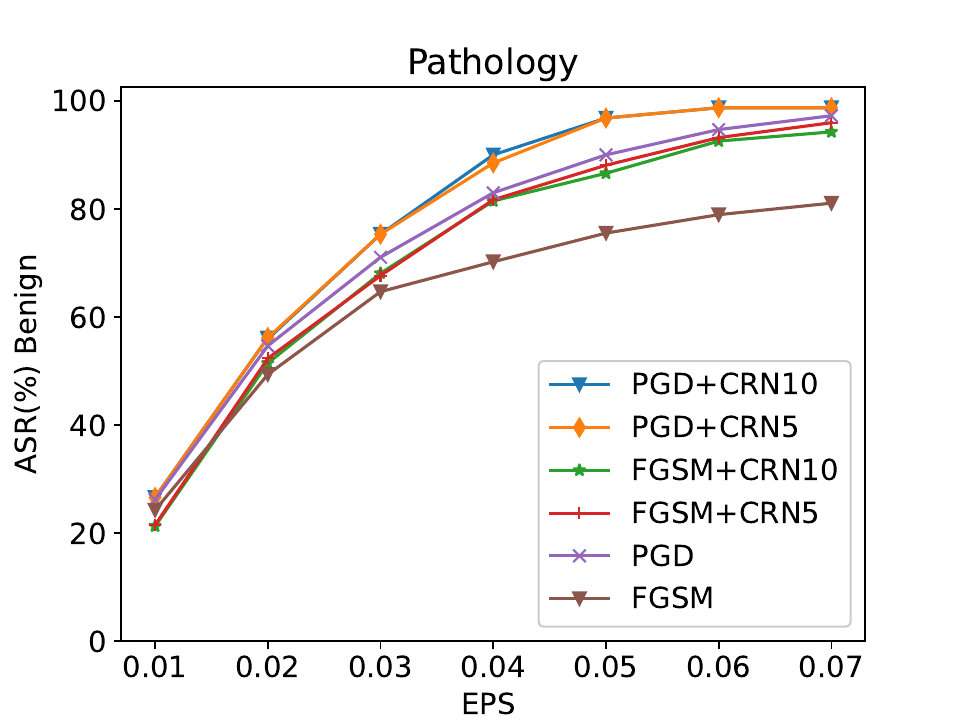}
         \caption{Effect of $\epsilon$ on Average ASR on benign clients }
         \label{fig:AASR benign-EPS}
     \end{subfigure}
     \hfill
     \begin{subfigure}[b]{1.05\textwidth}
          \centering
         \includegraphics[width=0.32\textwidth]{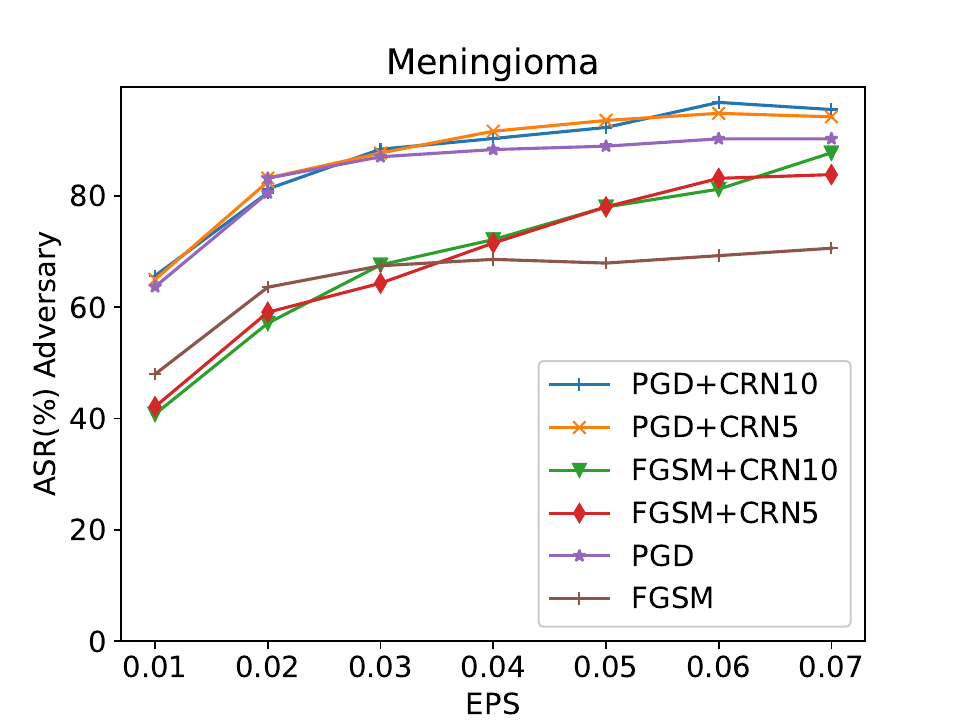}
          \includegraphics[width=0.32\textwidth]{Glioma_ASR_EPS-eps-converted-to.pdf}
           \includegraphics[width=0.32\textwidth]{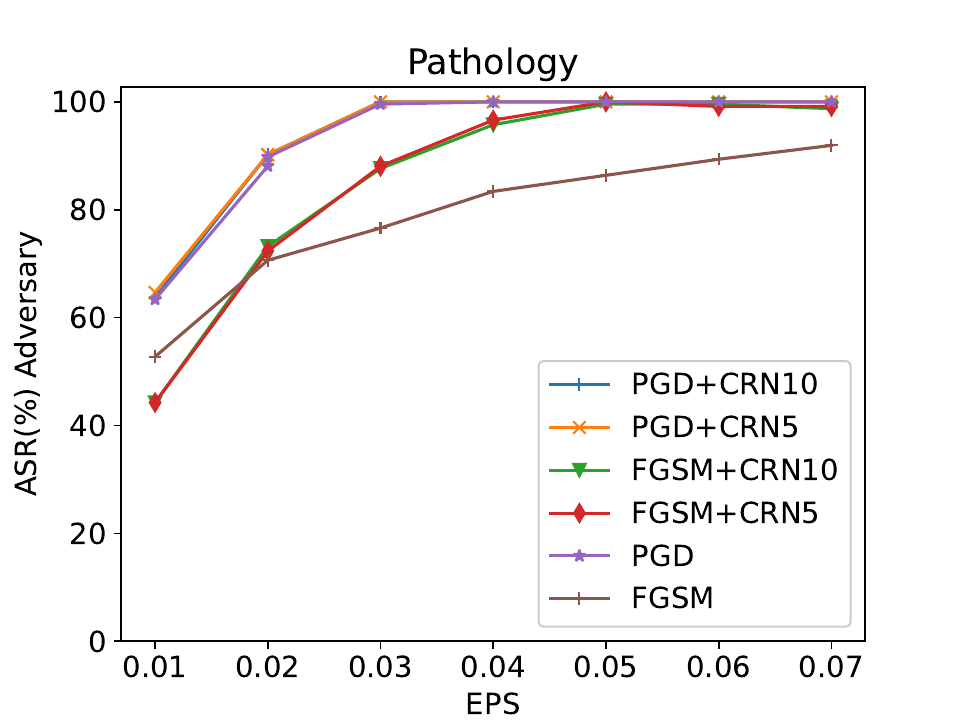}
         \caption{Effect of $\epsilon$ on ASR on the adversarial client }
         \label{fig:AASR malignant-EPS}
     \end{subfigure}
     \hfill
        \caption{Effect of Error perturbation degree $\epsilon$ on attack transferability. FGSM and PGD attacks with and without CRN initialization were performed.ASR is calculated on benign and adversary clients. The higher ASR on benign clients shows higher transferability}
        \label{fig: epsilon graphs}
\end{figure*}
\subsection{Efficiency Gains from CRN}

The role that CRN plays in enhancing the efficiency of adversarial attack strategies was explored by comparing the performance of the iterative attack, PGD, both in the presence and absence of CRN. Our primary metric was the time required to train an ensemble of adversarial samples. These evaluations were conducted under conditions characterized by \(\epsilon=0.05\) and \(\alpha=0.001\).

\begin{table*}[h]
\centering
\setlength{\tabcolsep}{8pt}
\renewcommand\arraystretch{1.4}
\caption{Result of Average Error Transfer Rate (AETR) for FGSM and PGD methods, with and without CRN initalization.} 
\begin{tabular}{| *{7}{c|} }\hline
 Dataset&\multicolumn{3}{c|}{ FGSM}     &              \multicolumn{3}{c|}{ PGD}                                                                           \\  

 \cline{2-7}
&Vanilla&CRN5 &CRN10&Vanilla&CRN5&CRN10\\\hline

Meningioma&84.80\%&\textbf{86.31}\%&{84.02}\%&83.73\%  &\textbf{84.25}\%  &84.16\%\\\hline
Glioma&82.32\% &87.06&\textbf{89.98\%}&74.89\%&\textbf{83.64\%}\% &82.32\%\\\hline

Pathology&\textbf{90.67\%}&{84.81}\%&{85.98}\%&79.86\% &86.79\%&\textbf{86.93\%}\\\hline

\end{tabular}
\label{AETR} 
\end{table*}

As detailed in Table \ref{table_time}, a key observation was that amplifying the number of iteration rounds had a twofold impact. Firstly, it increased the computational burden, necessitating more resources. Secondly, there was an increase in the AASR.  However, despite these increases, the cumulative computing time was substantially diminished compared to that required by conventional iterative models.

Incorporating CRN into the attack strategy was observed to improve transferability across disparate datasets. It's important to note that while computational complexity manifested a linear relationship with the iteration rounds, the enhancement in transferability was varying.

The PGD approach, especially when initialized randomly, exhibited a demand for computational resources. This trait poses potential bottlenecks when training with high-resolution images or for multiple image batches. Additionally, in situations where PGD's initial performance metrics are not high, a mere amplification in the number of iterations does not translate to significant performance increase.

The choice of initialization methodology appeared to be important in determining computational efficiency. The introduction of CRN as an initialization tool reduced the computational overhead. When compared against 40-iteration PGD, single-step adversarial attacks, when initialized with CRN, showed an improved efficiency, making them 20 to 30 times faster. An interesting observation was that, FGSM, when combined with CRN has higher performance than of PGD initialized via random methods.

An implication of the reduced GPU dependency is the bonus it offers to adversaries. By harnessing the power of CRN, adversaries can feasibly launch attacks on large data repositories or even mobile devices in an attack strategy. Traditional attack methodologies, which have been the benchmark in simulating adversarial attacks - using algorithms such as adaptive models and sanity checks as elaborated in \cite{carlini2019evaluating}, might require reconsideration. There's a possibility that adversaries, having more knowledge about the network, might deploy more efficient attack strategies or even use unconventional computational devices to perform the attack.

\subsection{Effect of $\epsilon$ on CRN-Enabled Attack Success}
We examined how the perturbation degree, $\epsilon$, influences the success rate of attacks on a given model. Our primary objective was to comprehend the interplay between the magnitude of perturbation and the attack's effectiveness. For this purpose, we set up various attack scenarios and compared two main categories, Baseline attacks and Attacks enhanced by the CRN.
Our results are quantitatively captured through the ASR. This rate was calculated for both benign and adversarial clients, and the detailed outcomes have been illustrated in Fig. \ref{fig: epsilon graphs}. From our observations, the following key insights emerged:

\textbf{Transferability and Attack Success}: Successful attacks executed on adversarial models tend to show an increased level of transferability to benign clients. This means that an attack that can efficiently compromise an adversarial model is also likely to be effective against benign ones.

\textbf{Effect of Low $\epsilon$ Values}: Perturbations with a smaller magnitude, especially when $\epsilon$ is as low as 0.01, result in noticeably diminished ASR rates across all client types. This suggests that minute perturbations are often insufficient to produce a successful attack.

\textbf{Trend with Increasing $\epsilon$ Values}: As the value of $\epsilon$ escalates, there is a consistent rise in the ASR. In simpler terms, a greater perturbation magnitude generally amplifies the probability of a successful attack.

\textbf{Performance on Pathology Images}: When assessing pathology images, iterative models exhibit high performance. They not only achieve high attack success rates against adversarial clients but also display commendable transferability.  Across all tested scenarios, the introduction of CRN consistently enhances the transferability of attacks. 

\textbf{Dependency of CRN-Enabled Models on $\epsilon$}: The impact of $\epsilon$ on CRN-enabled models is also noticeable. The extent to which this perturbation degree influences these models can differ based on the dataset in question, the chosen attack method, and the specific value of $\epsilon$.

\subsection{Error Transferability}

To gain insights into the transferability dynamics of adversarial attacks, we conceived a scenario, where an adversary cherry-picks only those samples from its subset that managed to successfully deceive its own model. These deceptive samples were then used to launch attacks on other clients models.

Our approach involved using two attack methodologies: Vanilla FGSM and PGD. Furthermore, we  integrated FGSM, and PGD initialized with CRN at two different configurations: CRN5 and CRN10. Our results are presented in Table \ref{AETR}.
The PGD method, in terms of raw attack success rate, was observed to surpass FGSM. However, when it came to AETR values, both PGD and FGSM were in close quarters, showing comparable performance.
    
\textbf{Role of Training Iterations in PGD:} For the PGD method, with 40 iterations, we evaluated its efficacy both in the presence and absence of CRN. Both FGSM and PGD, despite their differences, exhibited higher AETR values. This suggests that even if FGSM might, on average, generate adversarial examples that are less optimal, the few instances where it does succeed in deception are highly transferable to benign clients.
    
\textbf{Integration of CRN:} Incorporating CRN into the adversarial generation process was observed to boost the AETR values in our experiments. CRN simply increases transferability.
    
\textbf{Influence of $\epsilon$ on Transferability:} The perturbation magnitude, $\epsilon$, has positive effect on transferability  However, after crossing a certain perturbation threshold, mere amplification of $\epsilon$ does not correspondingly enhance the ASR. However, introducing CRN into this mix escalates this $\epsilon$ dependency. Through experimentation, we discerned an optimal perturbation range, which lay between 0.03 and 0.05.
    
\textbf{Centralized vs Federated Settings:} A key takeaway from our evaluations was the realization that conclusions derived in a centralized setting in previous research might not seamlessly extrapolate to a federated environment. This implies potential disparities in transferability trends or dependencies on specific parameters between these two settings.





\section{CONCLUSION}
In this study, we undertook a comprehensive investigation into adversarial attacks within a federated learning environment, using the information from previous global updates in attack strategies. Our findings indicate that CRN enhances the efficiency and transferability of adversarial attacks, especially when paired with iterative methods like PGD. By employing CRN as an initialization tool, adversaries can achieve enhanced computational efficiency, making attacks feasible on large datasets or on edge devices without high GPU support. A deeper dive into error transferability reveals that adversarial examples crafted through certain techniques, especially when combined with CRN, possess the capability to deceive a plethora of models. We found that adversarial dynamics prevalent in centralized settings do not necessarily translate directly to federated environments. This observation underscores the need for distinct considerations when transitioning to federated models, given the unique challenges they present. In summary, as federated learning continues to grow in popularity, understanding the intricacies of adversarial attacks in such settings becomes crucial. The results from this research not only provide a clearer picture of the current landscape but also serve as a foundation for the development of robust defensive strategies in future federated learning applications.

\bibliographystyle{IEEEtran}
\bibliography{IEEEabrv}


\vfill

\end{document}